\documentclass[runningheads]{llncs}
\usepackage{amsmath}

\usepackage{wrapfig, booktabs}
\usepackage{amsfonts}
\usepackage{bbm}
\usepackage{graphicx}
\usepackage{subfigure}
\usepackage{amsmath,amssymb}
\usepackage{times}
\usepackage{epsfig}
\usepackage{graphicx}
\usepackage{amsmath}
\usepackage{amssymb}
\usepackage{dirtytalk}
\usepackage{xcolor}
\usepackage{cite}
\usepackage{graphicx}
\usepackage{subfigure}

\usepackage{amsmath, amsfonts}

\DeclareUnicodeCharacter{2212}{-}
\begin{document}
\title{Leveraging Unsupervised Image Registration for  Discovery of Landmark Shape Descriptor}
%

%
\author{Riddhish Bhalodia  \and Shireen Elhabian \and Ladislav Kavan \and Ross Whitaker}

     
%
\institute{School of Computing, University of Utah, UT, USA }

\maketitle              
\begin{abstract}
In current biological and medical research, statistical shape modeling (SSM) provides an essential framework for the characterization of anatomy/morphology. Such analysis is often driven by the identification of a relatively small number of geometrically consistent features found across the samples of a population. These features can subsequently provide information about the population shape variation. Dense correspondence models can provide ease of computation and yield an interpretable low-dimensional shape descriptor when followed by dimensionality reduction. However, automatic methods for obtaining such correspondences usually require image segmentation followed by significant preprocessing, which is taxing in terms of both computation as well as human resources. In many cases, the segmentation and subsequent processing require manual guidance and anatomy specific domain expertise. This paper proposes a self-supervised deep learning approach for discovering landmarks from images that can directly be used as a shape descriptor for subsequent analysis. We use landmark-driven image registration as the primary task to force the neural network to discover landmarks that register the images well. We also propose a regularization term that allows for robust optimization of the neural network and ensures that the landmarks uniformly span the image domain. The proposed method circumvents segmentation and preprocessing and directly produces a usable shape descriptor using just 2D or 3D images. In addition, we also propose two variants on the training loss function that allows for prior shape information to be integrated into the model. We apply this framework on several 2D and 3D datasets to obtain their shape descriptors. We analyze these shape descriptors in their efficacy of capturing shape information by performing different shape-driven applications depending on the data  ranging from shape clustering to severity prediction to outcome diagnosis. The implementation is available at {\color{blue}\url{https://github.com/riddhishb/self-supervised-landmarks}}

\end{abstract}
\section{Introduction}
\label{sec:intro}

Statistical shape modeling (SSM) is an indispensable tool for the analysis of anatomy and biological structures. Such models can be viewed as a composite of two distinct steps: shape representation and shape analysis.  Shape representation is a quantifiable description of the shape/structure of sample from a population of anatomies (usually given as a cohort of images or surface meshes) that is consistent with the population statistics and is easy to use for subsequent analysis. There are two prominent families of algorithms for shape representation, (i) landmarks, which express shapes as point clouds that define an explicit correspondence map from one shape to another using invariant points across populations that vary in their form, and (ii) deformation fields, which rely on transformations between images to encode implicit shape information. Shape analysis then uses these shape representations to analyze the population's statistics; in most cases, the representation is projected onto a low-dimensional space via principal component analysis (PCA). This low-dimensional representation is used as a \emph{shape descriptor} for subsequent shape analysis.  Outside of analysis of different modes of shape variations captured by this descriptor, it can also be subsequently utilized in different applications. For instance, the shape descriptor can serve as features to perform classification of different morphological classes \cite{hufnagel2007shape}, can quantify the severity of a particular deformity \cite{bhalodia2019severity}, or employed to interpret and discover shape characteristics that are associated with a particular disease \cite{cates2014computational}. We consider such \emph{downstream applications} that are dependent on how well the shape descriptors characterize the given shape to showcase the efficacy of the shape descriptor.

Due to their simplicity and computational efficiency, correspondence-based models are the most prominently used models for shape representation. \emph{Correspondences} is a  term used to describe landmarks on the anatomy that are geometrically consistent across the samples of the population. In the earliest works, \cite{RTW:Tho17} correspondence was achieved by handpicked landmarks corresponding to distinguishable features. The field has come a long way with many state-of-the-art correspondence discovery algorithms \cite{styner2006spharm, cates2007shape}. However, many of these algorithms require segmentation of the anatomy from images as well as heavy pre-processing. Such segmentation and or pre-processing often come with a significant computational overhead as well as cost human resources. Segmentation of some anatomies is prone to subjective decisions and hence requires domain expertise. These problems fail to make the automated correspondence discovery model fully end-to-end, i.e., an automated pipeline that for inference just inputs images to produce shape descriptors for analysis.

In recent years, deep learning and neural networks models have had a significant impact on both image registration and shape analysis. With their ability to learn complex functions, several methods \cite{Bhalodia2018DeepSSM, milletari2017stats} have proposed learning correspondence from images, bypassing the need for segmentation and preprocessing. However, these methods are supervised and are data-hungry, they require considerable training data with correspondences, which is not always possible in clinical applications. They also need anatomy segmentation and preprocessing for the training set that might not be readily available. Deep networks have also played an essential role in developing computationally fast and unsupervised learning-based algorithms for image registration (e.g., \cite{balakrishnan2019tmi}) that perform equivalently to the state-of-the-art, optimization-based registration methods. However, transformations are not as friendly as correspondences for shape analysis; they often require the development of a fixed atlas \cite{RTW:Jos2004}. The systems that process image-to-image transformations express shape information in a high-dimensional space. Typically for shape analysis, a low-dimensional space is preferred, and therefore, these representations are projected onto a low-dimensional space via PCA (or some equivalent for nonlinear spaces), and the modes of shape variation need to be analyzed by domain experts to check for their usability in downstream applications. 

To address the above-stated challenges, we propose an end-to-end system for extracting a shape descriptor from only a population of input images. Ideally, this shape descriptor would not require any post-processing for subsequent analysis. This paper proposes a self-supervised deep learning approach for landmark discovery that uses image registration as the primary task. The proposed method alleviates the need for segmentation and heavy preprocessing (even during model training) to obtain a landmark-based shape descriptor. The discovered landmarks are relatively low in number; hence, they can be directly used for shape analysis and bypass the post-processing required to convert the representation into a low-dimensional space. The work presented here is an extension of the preliminary work presented in \cite{bhalodia2020selfsupervised}. This work significantly extends and improves on the previous paper in the following ways:
\begin{itemize}
    \item Additional experiments, results, and analysis on several different datasets with associated downstream applications for shape descriptors.
    \item We propose two different model variants that can incorporate prior information about shape into the model during training and can implicitly enforce the landmarks to encode such information.
    \item We propose an additional image matching loss function that preserves the local structure and allows for cross-modality registration or usage of datasets with a lot of intensity variations. 
\end{itemize}

\section{Related Work}
\label{sec:lit}

Since the groundbreaking work of D'Arcy Thompson \cite{RTW:Tho17} who utilized manually placed landmarks to study variations in shapes of fishes, statistical shape modeling (SSM) has become an indispensable tool for medical researchers and biologists. SSM finds applications in various fields such as cardiology \cite{cates2013afib}, neurology \cite{greig2001brain}, growth modeling \cite{RTW:Dat2009}, orthopaedics \cite{harris2013cam}, and instrument design \cite{goparaju2018evaluation}. Shape representation for SSM can be achieved via explicit representation of points on surfaces \cite{davies2002MDL, RTW:Sty2000}, direct usage of surface meshes or distance transforms \cite{mendoza2014personalized} or their features \cite{bouix2005hippocampal}, or, implicitly via functional maps \cite{ovsjanikov2012functional} or deformation fields \cite{beg2005computing}.

Correspondence-based models, or particle distribution models (PDMs) \cite{RTW:Gre91} place a dense set of particles onto the shapes' surfaces. Automatic PDM algorithms rely on non-linear optimization that reduces the complexity of the generative model \cite{cates2007shape, davies2002MDL}. In most cases, PCA is used to project the high dimensional shape space to a low dimensional \emph{shape descriptor} \cite{RTW:Tho17, bhalodia2019severity}. Since these algorithms require heavy pre-processing/segmentation, deep learning has been used to learn correspondences directly from a population of 2D/3D images \cite{Bhalodia2018DeepSSM, milletari2017stats}. These methods being supervised still require pre-processing overhead during training  and also need large datasets/data-augmentation methods to learn effectively. Both these requirements are not available in many cases, especially with medical data.

Also relevant is the work on deformable registration of images that have been used as a tool for shape representation \cite{beg2005computing} and atlas building \cite{RTW:Jos2004}. The implicit deformation fields are hard to interpret. Therefore, having a fixed atlas to which each image in the population will be deformed helps make the fields standardized. Deep learning-based unsupervised registration (e.g., \cite{balakrishnan2019tmi}) has attracted a lot of attention in recent years. This unsupervised registration framework have been extended in modeling diffeomorphic registration \cite{dalca2018unsupervised}, constructing image atlas \cite{dalca2019learning} and leveraging empirical information about the shape population \cite{bhalodia2019coop}.

Another relevant body of works stems from computer vision literature that uses image alignment to obtain dense feature maps \cite{detone18superpoint, Rocco17geomatch}, in a similar vein to the widely popular scale-invariant feature transform (SIFT) \cite{lowe2004distinctive} features. Other works focus on utilizing convolutional neural networks (CNNs) to learn surface features and use them to obtain correspondence points \cite{bronstein2016AnCNN} or as shape features for subsequent correspondence optimization \cite{agrawal2017deepfeatures}. All these works concentrate on discovering the surface/shape features using CNNs, whereas our work proposes an unsupervised approach for landmark discovery.

\section{Methods}
\label{sec:methods}

This section covers the necessary background for statistical shape modeling and image registration, the proposed model architecture and training, loss functions and optimization, and generalized model variants.

\subsection{Shape Analysis and Image Registration}
\label{sec:reg}
Statistical shape modeling (SSM) can be broadly categorized into two parts (i) shape representation and (ii) shape analysis. Shape representation entails using the raw data (can be in the form of images, meshes, label maps, etc.) and expressing it in a usable, quantifiable form for subsequent shape analysis. Shape analysis then finds relevant statistics from the shape representation pertinent to the downstream application.  To reiterate, in this paper, we refer to \emph{downstream applications} as the applications that utilize the shape descriptors to perform a task on given data, for example, using shape descriptors as features for shape classification. In this paper, we restrict our shape representation to be in the form of \emph{point correspondences}, which are a geometrically consistent set of 2D/3D points of the population of shapes. Hence, each shape $S_i$ from a population of shapes $S_1, ..., S_N$ can be expressed via $\mathcal{C}_i \in \mathbb{R}^{M \times d}$, where $M$ is the number of landmarks/correspondences per shape and $d$ is the space dimension ( $d = 2, 3$ for 2D and 3D shapes, respectively). We can use these $\mathcal{C}_i$'s for shape analysis, which usually involves performing PCA and using the low-dimensional representation for analysis of shape modes of variation. 

Landmarks also play an important role in image registration. A common underlying assumption in image registration is that a well-registered image will also match the landmarks placed at anatomically relevant features. Furthermore, landmarks can be used to perform image registration; for instance, radial basis functions (RBF) can be used to parametrize the image deformation based on landmarks (used as control points) provided on the source and the target images. Mathematically, if $T$ represents the image transformation, we can model the deformation as follows:
\begin{equation}
    T(\mathbf{x}) = \sum\limits_{i=1}^M w_i \phi(||\mathbf{x} - \mathbf{x}_i||)  + \alpha_1 \mathbf{x} + \alpha_0
    \label{eq:transform}
\end{equation}
Here, the $\phi$ represents the RBF function used, $\mathbf{x}$ represents the coordinates of the image, and, $\mathbf{x}_i$ represents the control points. If we are given the control points on source and target images, we can solve the linear system of equations to find $\mathbf{w} = [w_1, ..., w_M, \alpha_0, \alpha_1]$ and can apply the transformation to the entire image coordinate grid. The transformed coordinates is interpolated to obtain the warped image from the source image.

\begin{figure}[!h]
    \centering
    \includegraphics[width=\linewidth]{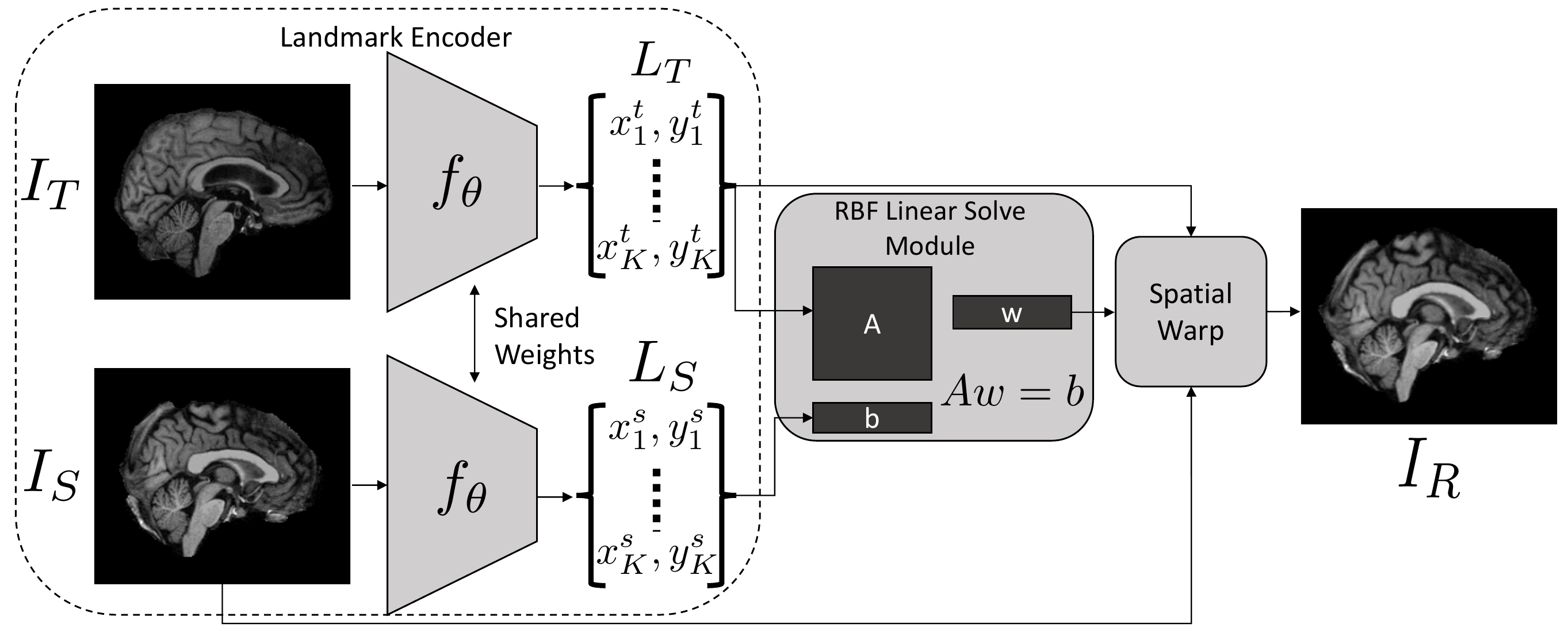}
    \caption{\textbf{Network Architecture}}
    \label{fig:model}
\end{figure}

\subsection{Model Description}
\label{sec:model}

Here, we propose a model to obtain anatomically relevant landmarks directly from images. To achieve this, we rely on the assumption that \textit{a good image registration between a pair of images should indicate good anatomical feature correspondence}. This assumption perfectly ties in with self-supervised learning, where we use image registration as a primary task and, in turn, obtain key landmark locations on the input images. 

The resulting architecture is broken into three components, broadly described as follows:
\begin{enumerate}
    \item[$-$] \textbf{Landmark Encoder: } This is a CNN network that operates on an image and outputs a $M\times d$ vector of landmark points. Since we work with the source (${I}_S$) and target (${I}_T$) images, landmark encoder $f_\theta$ is used twice, but share weights similar to a Siamese architecture \cite{koch2015siamese}.
    \item[$-$] \textbf{RBF Linear Solver: } We use the landmarks to construct a linear matrix $A = A({L}_T)$ and its associated output vector $b = b({L}_S)$. For clarification, we note here that the landmarks discovered by the landmark encoder function as control points for the RBF-based registration. For the transformation parameters $\mathbf{w}$, we have $A\mathbf{w} = b$. The RBF linear solver module consists of formulating this system of equations and solving them to find the transformation parameters. The matrix construction and linear solver are described in \ref{app:rbf}.
    \item[$-$] \textbf{Spatial Warp Module: }We use the transformation parameters and interpolate the source image and obtain the registered image ${I}_R$. This can be easily performed using a spatial transform unit \cite{jaderberg2015spatial}. 
\end{enumerate}

A detailed description of the architecture with layer description is given in \ref{app:arch}.
Throughout this work, we perform all of our experiments using thin-plate splines (TPS) as the kernel basis function, i.e., $\phi(r) = r^2 \log(r)$. The network architecture is described in Figure \ref{fig:model}.

\subsection{Loss Function and Regularization}
\label{sec:loss}



The training loss function of the proposed network can be given as follows:
\begin{equation}
    \mathcal{L} = \mathcal{L}_{\text{match}}({I}_T, {I}_R) + \lambda \mathcal{L}_{\text{reg}}(A)
    \label{eq:main-loss}
\end{equation}
The first term is the \emph{image matching} or the registration loss between the target image and the registered (i.e., warped source) image. The second term is the regularization of the registration system applied on the matrix $A$ from the RBF linear solve module. We shall describe both these terms in detail in  the following paragraphs.

\subsubsection{Image Matching Loss}
\label{sec:match}

In classical literature for image registration, there is a research emphasis on interpretation as well as effect of utilizing different loss functions in context of image matching optimization \cite{tagare2014does}. In the context of deep learning, the methods are restricted to loss functions that allow for back-propagation. The two most commonly used image-matching functions used in deep leaning based image registration are the $\mathbb{L}$2 and the normalized cross-correlation (NCC) loss. These are given as follows:
\begin{align}
    \mathcal{L}_{\text{match-L2}} &= ||I_T - I_R||^2 \\
    \mathcal{L}_{\text{match-NCC}} &= 1 - \frac{1}{|\mathcal{P}|} \sum_{\mathbf{x} \in \mathcal{P}}NCC(I_T(\mathbf{x}, p), I_R(\mathbf{x}, p))
\end{align}
Here,  $I(\mathbf{x}, p)$ represents an image patch on image $I$ centered at voxel location $\mathbf{x}$ with patch size of $p$. $\mathcal{P}$ denotes the set of all patches/possible patch centers. The function $NCC(I_T(\mathbf{x}, p), I_R(\mathbf{x}, p))$ represents the normalized cross-correlation between two patches.

Both the $\mathbb{L}$2 loss and NCC work on the pixel intensities and work well when the data population has a consistent intensity profile across the population.  CT scans are a good example of images with a consistent intensity profile. However, in datasets such as cardiac MRI, the intensity histograms of individual MRI scans are highly variable, and these intensity-driven loss functions will fail to capture structural matching. These losses will also fail when the input data comes from two different sources or modalities, such as two different scanners/centers or a dataset containing both T1 weighted MRI and T2 weighted MRI images. In such scenarios, pixel intensities are not the correct measure to quantify image matching, and we need losses that can capture structural correlation. Therefore, we also use the modality independent neighborhood descriptor (MIND) features to formulate a registration loss; several other recent registration works have used MIND features as loss \cite{xu2020adversarial}. MIND features rely on image patches; for a given image $I$ at a pixel/voxel location $\mathbf{x}$, its image patch is denoted as $I(\mathbf{x}, p)$, with $p$ being the patch size. The MIND feature for an image at a pixel/voxel is given as:

\begin{align}
    MIND(I, \mathbf{x}, p, \mathbf{r}) = \exp(\frac{-||I(\mathbf{x}, p) -  I(\mathbf{x} + \mathbf{r}, p)||^2}{\text{Var}(I( \mathbf{x}, p))})
\end{align}

Here, $\mathbf{r}$ is the displacement vector and $\text{Var}(.)$ is the local variance of an image patch. The match loss function using these MIND features is given as:
\begin{align}
    \mathcal{L}_{\text{match-MIND}} = \frac{1}{|\Omega||\mathcal{R}|}\sum_{\mathbf{x} \in \Omega}\sum_{\mathbf{r} \in \mathcal{R}}|MIND(I_T, \mathbf{x}, p, \mathbf{r}) - MIND(I_R, \mathbf{x}, p, \mathbf{r})|
\end{align}

$\Omega$ is the set of voxel locations and $\mathcal{R}$ is the set of displacements used. We generally use a set of displacement vectors describing a local neighborhood (such as 4-neighbor for 2D images or 6-neighbor for 3D). The loss function is parametrized by (i) the patch size $p$, and (ii) the distance value, i.e. $||\mathbf{r}||$. In all of our experiments using MIND loss both on 2D and 3D, we use  an isotropic patch of size 3 and the displacement is kept as $||\mathbf{r}|| = 5$.

\subsubsection{Regularization}
\label{sec:reg}

\begin{figure}
    \centering
    \includegraphics[width=\linewidth]{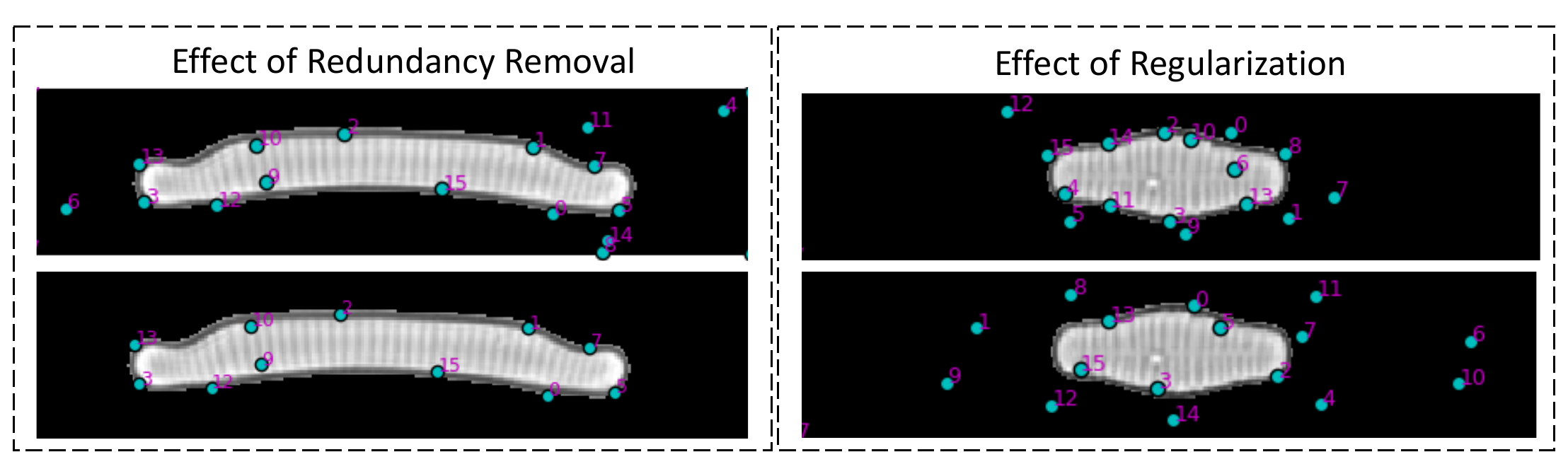}
    \caption{Effect of regularization (right), where the top image is un-regularized and the bottom one is regularized. Effect of redundancy removal (left), where top figure is before redundancy removal and bottom figure is after.}
    \label{fig:reg}
\end{figure}

The linear system required to solve the RBF warp parameters requires that matrix $A$ is a non-singular matrix. However, the positions of the landmarks coming from the landmark encoder are arbitrary. Hence, during the optimization, the matrix $A$ can be poorly conditioned or even singular. A singular matrix has an infinite condition number, and a poorly conditioned matrix has a large condition number. Such a scenario can result in infinite number of solutions to the linear system and can cause optimization to break. To ensure stable optimization, we introduce a regularization term that minimizes the condition number of the matrix $A$, given as:
\begin{align}
    \mathcal{L}_{\text{reg}} = \kappa(A) = ||A||_F ||A^{-1}||_F
\end{align}
$||.||_F$ denotes the Frobenius norm of the matrix; this allows us to easily differentiate the regularizer. Minimization of the condition number of $A$ provides an additional benefit. We note that a poorly conditioned $A$ occurs when one or more pair of landmarks (used for the construction of $A$) are very close together. Hence, making the matrix well-conditioned will force the landmarks to be spread out more throughout the image, allowing us to span larger regions. This effect is showcased in Figure \ref{fig:reg},  demonstrated on diatoms, which is a simple example. One can note that the landmarks spread on the diatom boundary are achieved even with no regularization. However, we see a lot of landmarks being close together that can cause the optimization to break. This is indeed the case as the optimization without regularization is observed to break (generate NaNs) because of the singularity of the kernel matrix $A$. With points close together, the overall shape descriptor is not informative. The regularized version spreads landmarks more uniformly over the image domain. The landmarks that are far from the boundary can be removed by redundancy removal described in the next section. The effect of regularization on the registration loss is described in the results section \ref{sec:diatom}.

\subsection{Training Procedure}
\label{sec:train}



The entire model is trained jointly with Adam \cite{kingma2014adam}, and network parameters are initialized randomly in showcased experiments. However, one can imagine a starting initialization for the output layer of the landmark encoder. One such initialization could be by using the mean landmarks from a pre-compute PDM on the population. The choice of hyper-parameter $\lambda$ controls the amount of regularization and can be chosen via cross-validation. For a given set of images, the training is performed on a set of image pairs. For smaller datasets, we use all possible image pairs, and for larger datasets, we can either randomly choose image pairs or employ a sampling heuristic. As the model is trained on pairs of images, even small datasets (as small as ~50 images) can be effectively used for training the model. Such a low-resource learning is imperative for medical images where data is scarce and cannot be effectively used to train neural networks.

In this work, we treat the number of landmarks to be predicted as a hyper-parameter. As the landmark locations can be arbitrary, there could be redundant landmarks in characterizing the structure of interest. To remove such landmarks, we use a simple heuristic,  based on the assumption that \textit{removing such redundant landmarks will not affect the registration}. Using this heuristic, we compute the change in mean registration accuracy by removing one landmark at a time (this is performed on an already trained model as post-processing). This difference is the importance value attached to each landmark, as more significant the difference more is the importance of the landmark in performing accurate registration.  We remove these landmarks in a greedy fashion, that is, we follow these steps.
\begin{enumerate}
    \item For remaining set of landmarks compute the registration loss. Initially the set of remaining landmarks is same as the starting set of landmarks.
    \item Temporarily remove one landmark at a time and compute the registration loss and its difference from one computed in step 1. Do this for all remaining landmarks.
    \item Select the landmark with least difference (least importance) and remove it.
    \item Repeat steps 1-3 till either desired number of landmarks are reached or a difference threshold is reached.
\end{enumerate}
This removal allows for a smaller and more informative landmark-based shape descriptor, and the effect is shown in Figure \ref{fig:reg}.

 \textbf{Regularization and redundancy removal: } The regularization term tends to spread the particles evenly across the image and is applied as a soft constraint with the image matching loss. The regularization acts in conjunction with the registration loss, i.e., if a feature in an image exhibits a higher registration loss, the particles will be distributed to match that feature better. In cases where the anatomy of interest is localized with lower registration loss in that region would cause the redundancy removal to disregard the spread-out particles. In such scenarios, the registration loss must be spatially-weighted to introduce a preference to the localized anatomy; this model variant is introduced in the following section. Furthermore, the redundancy removal needs to be applied carefully with quality control to remove particles from the region of interest. We can also modify the redundancy removal process to only look at selective regions in the image while computing the registration accuracy.

\textbf{Note on Training and Inference Time: } We trained the network (2D architecture described in Appendix \ref{app:arch}) with 30 2D landmarks on a dataset of 100 toy images of $256\times 256$ dimensions (that is 10000 pairs -- actual training size). We utilize a single 12GB NVIDIA TITAN V GPU. The training time for one epoch with batch size of 20 is 6.2 minutes, and the inference time for a single scan is 0.4 seconds. Another important aspect is the GPU memory requirement, which for this experiment is 4600MB utilizing Pytorch deep learning framework.

\subsection{Model Variants}
\label{sec:variants}

\textbf{Weak Supervision Variant: }The proposed model can be easily modified for a weakly supervised learning framework to introduce a prior that informs the landmark positioning via changes to its loss function. In many cases, medical data can be presented along with some form of shape description. The most common of these forms is via segmentation of the anatomy of interest. However, in a typical scenario, only a limited number of data have segmentation associated with the image. In such cases, segmentation can be used to improve the landmark positioning by the following changes to the model:
\begin{itemize}
    \item[$-$] We use the learned transformation parameters to transform a source segmentation image ($C_S$) to the registered segmentation ($C_R$).
    \item[$-$] Introduce the matching function between the registered segmentation to the target segmentation ($C_T$) and optimize the model with the updated loss. This loss function will only be activated when both the source and target segmentations are present, providing weak supervision for the landmark (shape descriptor) discovery task.
\end{itemize}

The loss function can thus be expressed as:

\begin{align}
    \mathcal{L} = \mathcal{L}_{\text{match}}({I}_T, {I}_R) + \lambda \mathcal{L}_{\text{reg}}(A) + \beta \mathbbm{1}_{C} \mathcal{L}_{\text{match}}({C}_T, {C}_R)
    \label{eq:ws-loss}
\end{align}

Here, $\mathbbm{1}_C$ is an indicator variable that is 1 when both $C_T$ and $C_S$ exists and zero otherwise.

There are two other aspects to note here: (i) the input to the landmark encoder are still images, and therefore during testing, we do not need an additional segmentation input, and (ii) instead of binary segmentation, any other forms of shape information that can be deformed can be used as well, such as signed distance transforms or correspondences.

\textbf{Localized Variant: }  In some instances, the anatomical area of interest is localized, and we would like the landmarks to be expressive of that particular location. For instance,  in CT scans of the Femur, the diagnosis and characterization cam-type femoroacetabular impingement  \cite{Stephanie2015impingment} is done by analyzing the localized region below the femoral head and how its difference from  a representative femur shape or a healthy patient. Such localized characterization requires the shape descriptor to contain sufficient information about this localized anatomy of interest. To achieve this, we again assume that image registration enforces landmark location, i.e., for the landmarks to be more expressive of the localized region, we want to achieve the best image registration in that region. Therefore, we propose a simple modification to the loss function.

\begin{align}
    \mathcal{L} = \mathcal{L}_{\text{match}}(m_T \circ I_T, m_R \circ I_R) +  \lambda \mathcal{L}_{\text{reg}}(A)
    \label{eq:loc}
\end{align}

Here, $m_I, m_R$ represents a mask representing the location of the localized region of interest. In a special case, these masks are fixed (i.e., the same) for the given dataset if the anatomy of interest occupies a common space across the image population, i.e., images are roughly aligned.

\section{Results}
\label{sec:results}
This section shows the results of the proposed methods on different 2D/3D datasets and is divided into subsections corresponding to each dataset. We also demonstrate the usefulness of the landmark-based shape descriptor obtained in each case paired together with a downstream application. This section also includes an analysis of regularization, redundancy removal, and the application of different proposed framework variants. In most cases, the number of  epochs are chosen via early stopping based on the best validation loss. 

\subsection{Diatom Dataset}
\label{sec:diatom}

Diatoms are a biological group of algae found in different water bodies and fossilized deposits. Diatoms are unicellular and are categorized into different classes based on their shape/structure. Any characterization of diatoms is based on their size and shape; therefore, it is a perfect dataset for applying the proposed method and finding landmark-based shape descriptors. The diatoms dataset \footnote{Downloaded from the DIADIST project page: rbg-web2.rbge.org.uk/DIADIST/} contains 2D isolated cellular images from four different diatom morphologies, namely, Eunotia (68 samples), Fragilariforma (100 samples), Gomphonema (100 samples), and Stauroneis (72 samples). This dataset is collected as part of the automatic diatom identification and classification project \cite{du1999diatom}. The data is split into 80\%, 10\%, 10\% for training, validation, and testing datasets.

\begin{figure}
    \centering
    \includegraphics[width=\linewidth]{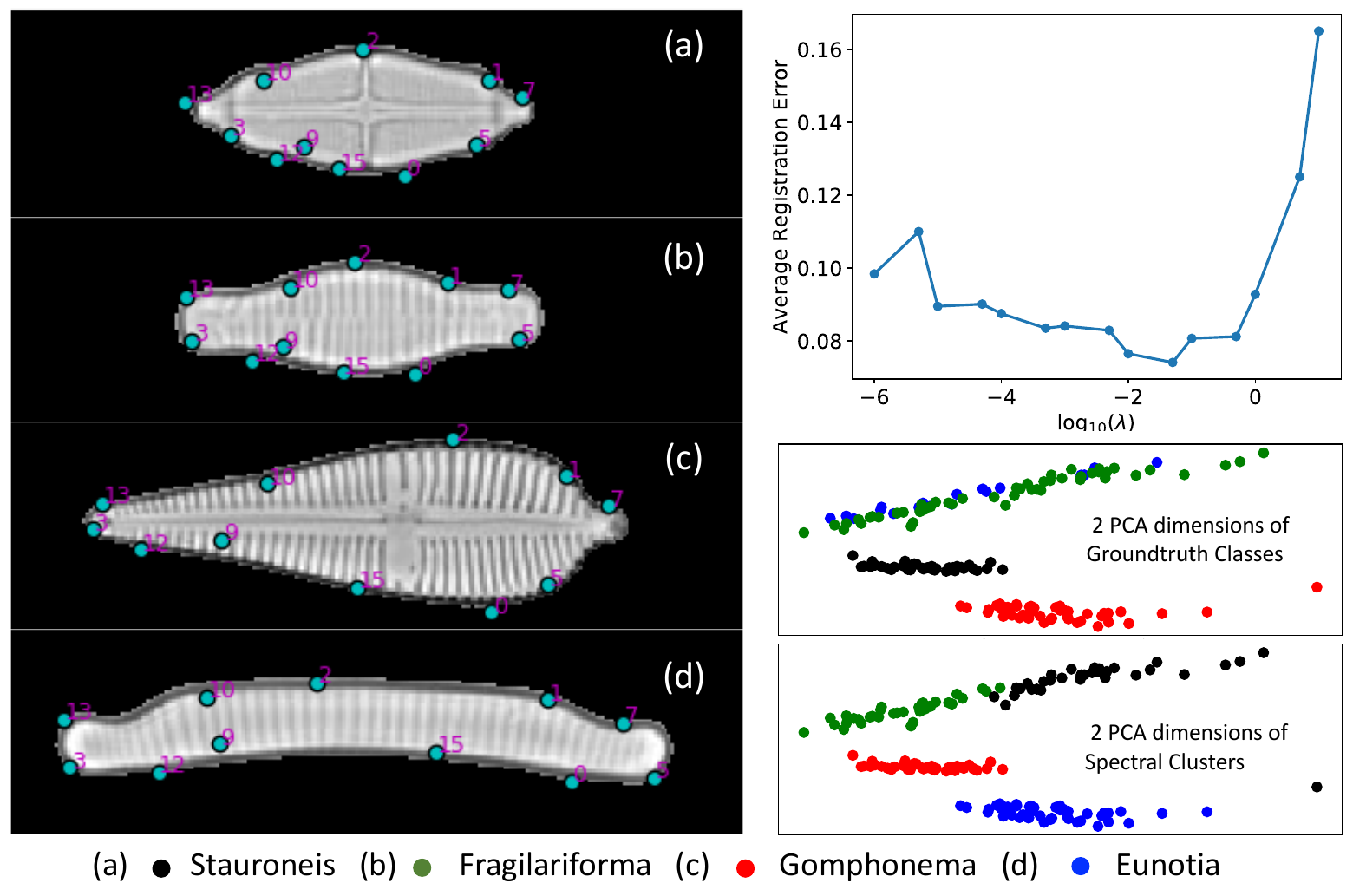}
    \caption{\textbf{Diatoms results } The left images show landmarks (after redundancy removal) on test images from four different classes and are in correspondence. The top right plot shows the effect of the regularization parameter $\lambda$ on the registration accuracy. The two scatter plots on the bottom right shows the results of performing unsupervised clustering using landmarks as features compared to ground truth labels.}
    \label{fig:diatoms}
\end{figure}

We train the proposed network with $\mathbb{L}$2 loss for image matching, with a regularization parameter of $\lambda = 0.005$ (found using cross-validation as described ahead), and with 16  landmarks. We also keep four pre-determined landmarks on the corners while computing the warp; these are not learned via the network. We train the network (using 2D image architecture as given in \ref{app:arch}) for 20 epochs on all possible image pairs, with no additional data augmentation. As a post-processing step, we perform the redundancy removal as described in Section \ref{sec:train} to retain 11 landmarks. Results shown in Figure \ref{fig:diatoms} highlight the structural correspondence between different diatoms classes.  We can notice that some of the landmarks are not precisely on the border of the shape. Such positioning of landmarks arises from the fully unsupervised training of the model with respect to the landmarks. The network has no prior on how and where to place the landmarks, and hence, the placement of the landmarks is the result of having the best possible registration loss. For instance, the landmark number 9 in Figure \ref{fig:diatoms} is dictated via the registration error that placing it inside the image border in that reasonably similar intensity level will reduce the image matching loss.

\textbf{Downstream Application: }These landmarks are easy to use as shape-based features of diatoms. We perform spectral clustering \cite{von2007tutorial} using these landmarks as features, and it performs well to separate the classes into clusters except for \emph{fragilariforma} and \emph{eunotia} that exhibit very similar shapes when their scales match. For classification, a multi-layer perceptron (with a single hidden layer) can distinguish between these four classes using these landmarks as inputs with 100\% test accuracy. 

\textbf{Regularization parameter: } Using this simple dataset, we also want to showcase the selection of regularization parameter $\lambda$ via cross-validation. We perform three-fold cross-validation with different lambdas and compute the average registration loss at every fold. The plot for this experiment is shown on the top-right of Figure \ref{fig:diatoms}.  It highlights that there is an optimal $\lambda$ that minimizes the registration loss and is a notable result. Since the regularization does not act on network parameters and limit the space of solutions, there is no guarantee that the generalization in registration performance should improve.
The regularization term is only introduced for optimization stability, ensuring uniform particle spread tasks that it performs well.  However, empirically, we see an optimal $\lambda$ that achieves the best generalization, showing that the proposed regularization term also improves the generalization of the network.

\subsection{Metopic Craniosynostosis Dataset}
\label{sec:cranio}

Metopic Craniosynostosis  \cite{david2011craniosynostosis} is a morphological condition of the skull/cranium that affects infants. It occurs due to premature fusion of the metopic suture in the skull, and the subsequent brain growth causes deformed head shapes.  The morphological symptoms include a triangular-shaped forehead (trigonocephaly) \cite{kellogg2012interfrontal} and compensatory expansion of the back of the head. In severe cases, along with abnormal morphology, patients are affected by the increased intracranial pressure causing several neurological complications. In current practice, the severity of metopic craniosynostosis is gauged subjectively by surgeons, affecting the  subsequent treatment protocol. The usual treatment entails a risky surgical procedure for severe cases or continued observation for milder ones. In recent research, the skull shape of metopic patients and its deviation from normal has been used for devising an objective severity measure \cite{kellogg2012interfrontal, Bhalodia2018DeepSSM}. These methods use CT scans that underwent segmentation and/or are processed for shape representation; these steps involve manual and computational overhead.

\begin{figure}[!h]
    \centering
    \includegraphics[width=\linewidth]{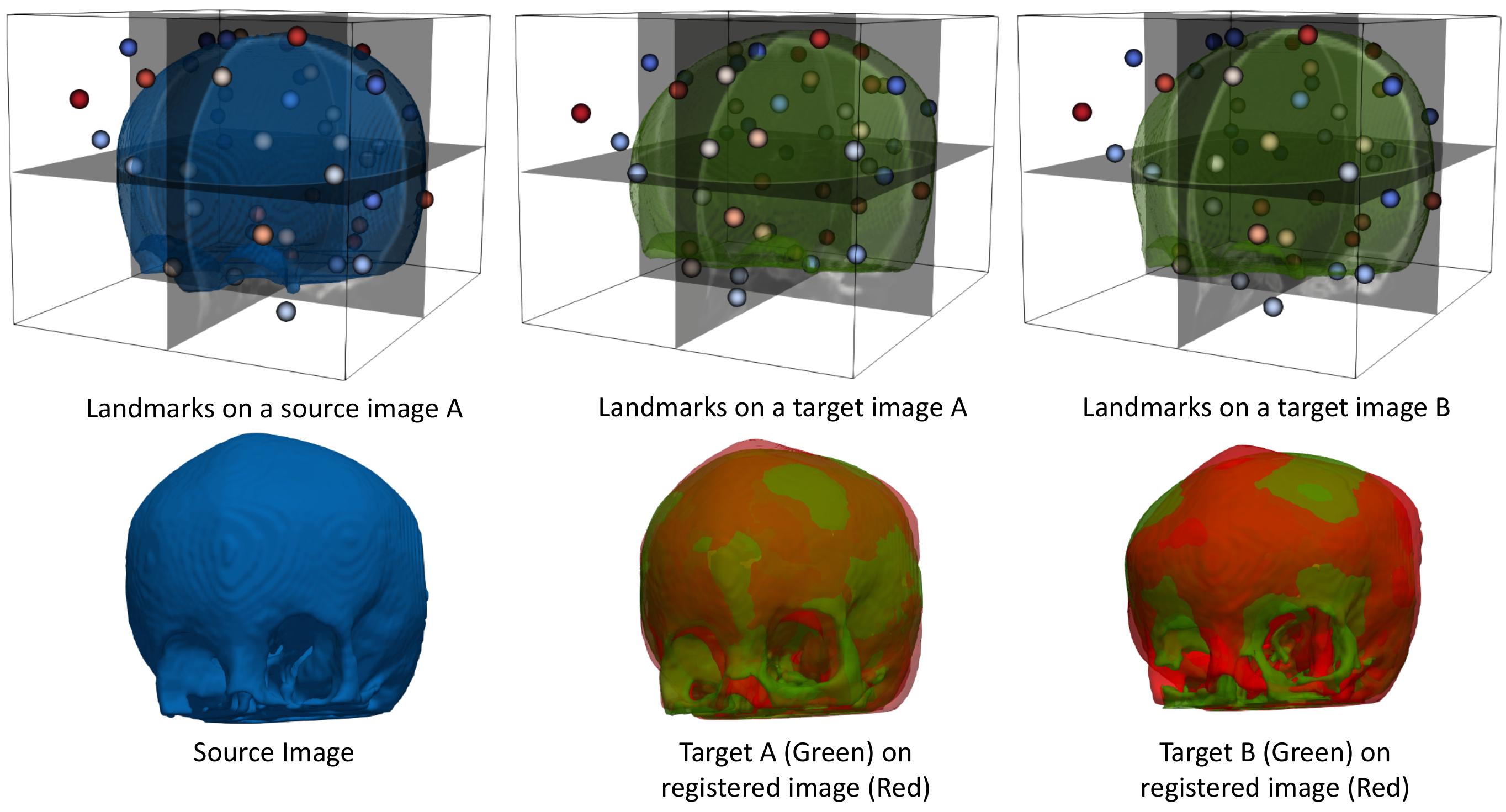}
    \caption{\textbf{Metopic craniosynostosis results } where we show a single source image (metopic) with two different target images (A:metopic, B:control). The top row showcases the landmark positions shown in 3D space with an overlay of segmentation mask and mix-axial,mid-sagittal, and mid-coronal slices. The bottom row showcases the registration via overlayed segmentation masks.}
    \label{fig:cranio}
\end{figure}

We use the proposed method directly on the CT scans and aim to obtain a shape descriptor that can be subsequently used for severity quantification of metopic craniosynostosis.  Our dataset comprises cranial CT scans of infants between 5-15 months of age, these scans were acquired at UPMC Children’s Hospital of Pittsburgh between 2002-2016. Out of which 27 are scans of patients diagnosed with metopic craniosynostosis, this diagnosis was performed by a board-certified craniofacial plastic surgeon utilizing the CT images as well as a physical examination. The remaining 93 patients were trauma patients that underwent CT scans; however, they demonstrated no morphological abnormalities, and these scans form our control set. All the CT scans were acquired as a part of routine clinical care protocol and under an IRB-approved registry. We collect this data with HIPAA protocols and de-identification of the scans. In the following analysis, we refer to the CT scans from the control set as \emph{controls} and the CT scans that are diagnosed with metopic craniosynostosis as \emph{metopic}.  
We train the network \emph{only on the control set} with 80\%, 10\%, 10\% data split for training, validation, and testing, respectively. We use $\mathbb{L}$2 as our image matching loss function,  because the dataset has minimal intensity variation across different samples, and the primary structure of the cranium is well-defined. We use 100 3D landmarks (8 constant on corners), and a regularization parameter of $\lambda = 0.00001$ (discovered via cross-validation). We perform redundant landmark removal by selecting the best 50 landmarks as post-processing. In addition to the test set, we also evaluate the model on the  CT scans diagnosed with metopic craniosynostosis, these CT scans are not observed by the model during training as it is only trained on the control set.  Figure \ref{fig:cranio} showcases qualitative results where we see that the registration performance between a single source image (metopic) and two targets, one metopic other from the control set. 

\textbf{Downstream Application: }Using the landmarks from the proposed method as a shape descriptor, we can compute the Mahalanobis distance/Z-scores of each scan with respect to a population of control scans.  If we consider $\mathbf{z}_1, ..., \mathbf{z}_N$ be the shape descriptors of N scans from the control set, whose mean and covariance is given as $\mathbf{\mu}$ and $\Sigma$ respectively, the Mahalanobis distance/Z-score for a data with shape descriptor $\mathbf{z}$ is given by:
\begin{align}
    ZS(\mathbf{z}) = \sqrt{(\mathbf{z} - \mathbf{\mu})^T\Sigma^{-1}(\mathbf{z} - \mathbf{\mu})}
\end{align}

\begin{figure}[!h]
    \centering
    \includegraphics[width=\linewidth]{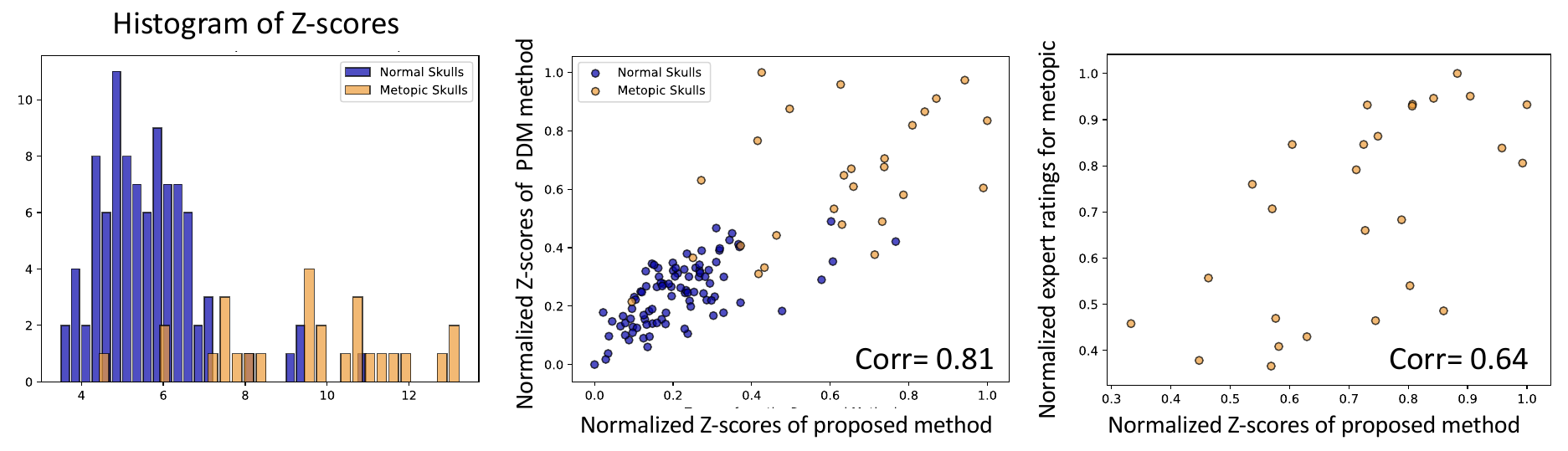}
    \caption{\textbf{Metopic severity analysis} the plot on the left shows the histogram of Z-scores using the landmarks as a shape descriptor. The figure in the middle shows the correlation with Z-score from the state-of-the-art correspondence model. The figure on the right shows the correlation between the Z-scores of the metopic scans and the aggregate rating of these scans given by experts. }
    \label{fig:cranio-scatter}
\end{figure}

This Z-score represents the deviation of a scan from the control population (this set does not have any symptoms of cranial deformity). Hence, a metopic skull would have a larger Z-score than a control scan. Figure \ref{fig:cranio-scatter} (left) shows that this is indeed the case.  We also want to compare the efficacy of the obtained landmarks as shape descriptors with respect to dense correspondences from a points distribution model(PDM). For this, we utilize \emph{ShapeWorks} \cite{cates2017shapeworks} package, a state-of-the-art method for automatic correspondence discovery. As a PDM method, \emph{ShapeWorks} has been used in the context of metopic severity prediction \cite{bhalodia2019severity}. ShapeWorks produces a set of 2048 correspondences on surfaces of cranial CT scans using their segmented images, which is represented as a low-dimensional PCA loading vector to be used as a shape descriptor. We compare the efficacy of both the shape descriptors in characterizing the severity of metopic craniosynostosis. For this, we compute the Mahalanobis Distance (or Z-score) of each shape (landmark shape descriptor) with respect to control data distribution. The Pearson's correlation between two scores is 0.81, and its scatter plot (after normalizing each set of Z-scores) is given in Figure \ref{fig:cranio-scatter}. The Z-scores are normalized by dividing with the maximum Z-score from the set, this allows them to lie between 0 and 1 and provide better visualization in the scatter plot, this normalization does not affect the correlation score. Such a significant correlation showcases that both methods capture similar shape information required to characterize the severity of the condition.  Additionally, we compare the Z-scores from the proposed method with aggregate severity scores from 16 craniosynostosis experts' ratings. Each rating uses a Likert scale between 0-5, with 5 being the most severe, and only 27 metopic scans are rated. The Z-scores and the aggregate ratings show a positive correlation with Pearson's coefficient of 0.64 (see Figure \ref{fig:cranio-scatter}). In comparison, Pearson's correlation between the Mahalanobis distance from ShapeWorks and the expert ratings is only 0.28.

\subsection{Cam-type Femoroacetabular Impingement (cam-FAI)}
\label{sec:femur}

\begin{figure}
    \centering
    \includegraphics[width=0.9\linewidth]{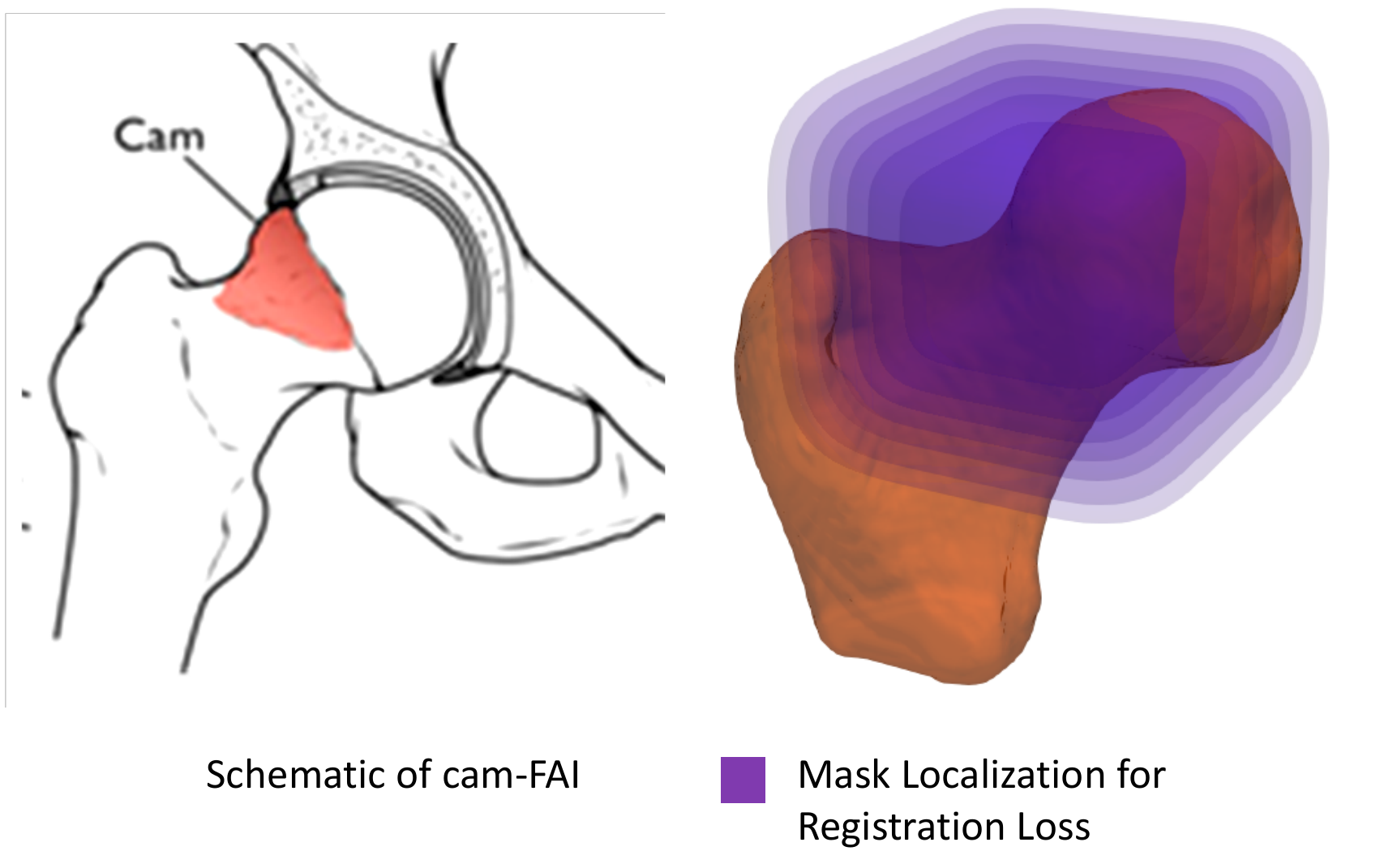}
    \caption{\textbf{Cam-FAI} The left figure shows location of cam-FAI and the right figure showcases the location of mask onto a median femur anatomy denoting location of interest to gauge shape descriptor of pathology.}
    \label{fig:cam_descp}
\end{figure}
\begin{figure}
    \centering
    \includegraphics[width=0.85\linewidth]{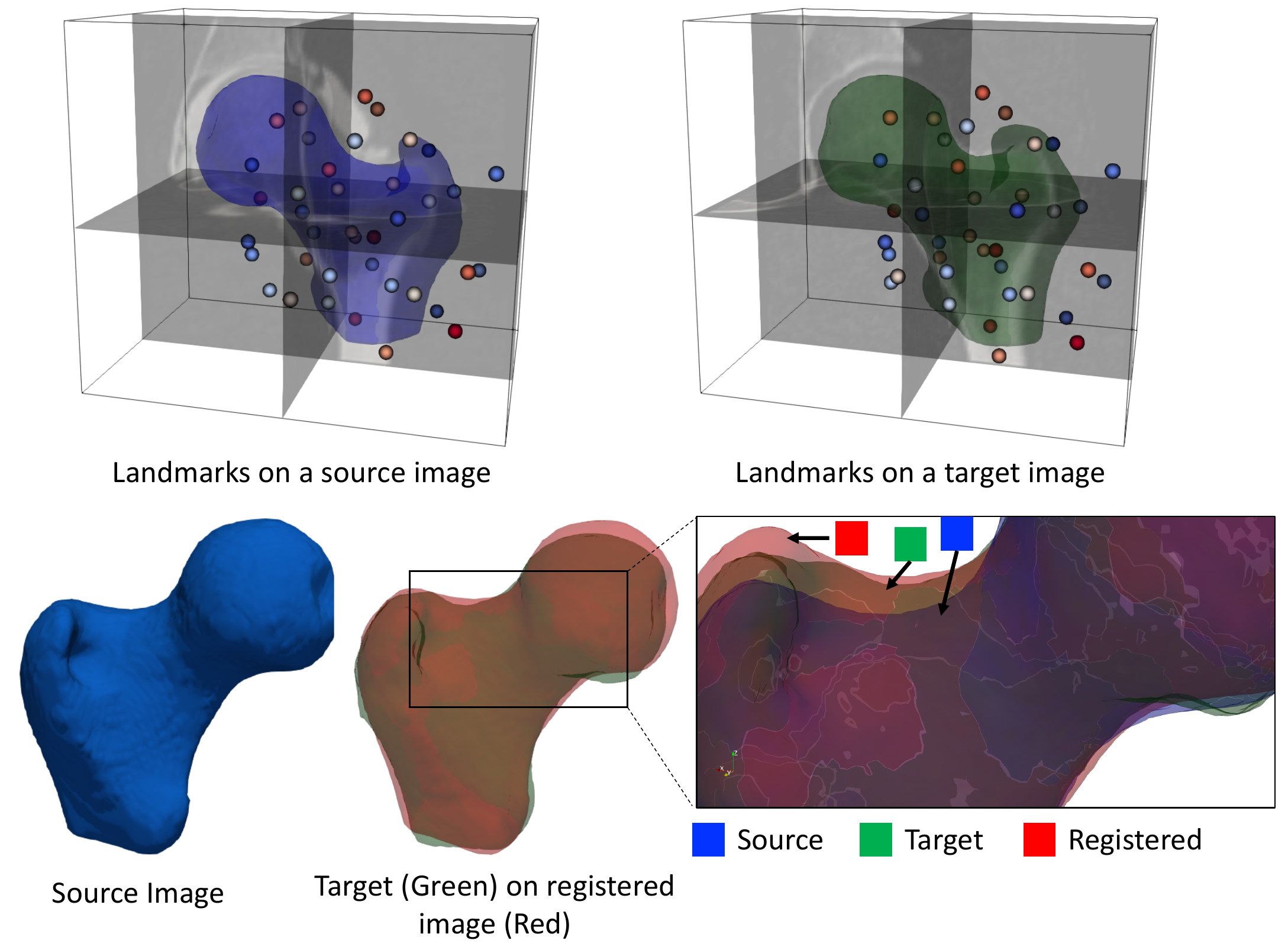}
    \caption{\textbf{Femur results} where we show landmarks discovery and registration on a single source image (a Cam-FAI diagnosed subject) with and target image (normal). The bottom right shows zoomed-in version of the registration error near the location of interest.}
    \label{fig:femur_results}
\end{figure}

Femoroarticular impingement (FAI) \cite{atkins2017quantitative} is an orthopedic disorder that affects movement in the hip joint.  Cam-FAI is a primary
cause of hip osteoarthritis and is characterized by an abnormal bone growth of the
the femoral head. Cam-FAI affects a localized region as shown in Figure \ref{fig:cam_descp}. The deviation of a  femur anatomy diagnosed with Cam-FAI to representative femur anatomy from a healthy patient population is of interest. This deviation can inform operative decisions and subsequent treatment planning. With the localized variant of the proposed method (see Section \ref{sec:train}), we aim to discover the shape descriptor that captures the localized structure of the cam-FAI pathology.  In this study, we use a dataset of 59 CT scans of the femur bone, out of which 50 scans are of patients without any diagnosed morphological defect in their femurs, we call this the control set. Additionally, we also have another 9 CT scans of the femur bone that are from patients diagnosed with Cam-FAI deformity. All data was originally collected for research purposes,  and specifically for the evaluation of hip bio-mechanics \cite{harris2013cam, harris2013three} at Orthopaedics Research Laboratory, School of Medicine, University of Utah. All participants provided informed consent prior to participation in this University of Utah IRB-approved study. The data contains femur scans of both left and right femur, and all the right femur have been reflected from the mid-saggital plane to have consistent orientation across the dataset. We use the median CT scan of the control set to define a common/fixed mask image for the image matching loss (Eq \ref{eq:loc}). This is defined by selecting a bounding box around the anatomy of interest and then blurring it using a convolution with a Gaussian filter ( Figure \ref{fig:cam_descp}). We apply the method with $\lambda=0.000001$ (found by cross-validation) and train the model for 20 epochs on the  joint set (controls and cam-FAI diagnosed scans) dataset with 80 landmarks, and perform redundancy removal by selecting the top 40 landmarks. Figure \ref{fig:femur_results} showcases the landmarks on a single source-target image pair and their corresponding registration output. We notice that the registration error is low in the region of interest, as stressed by the closeup section of the registration. 

\begin{figure}
    \centering
    \includegraphics[width=\linewidth]{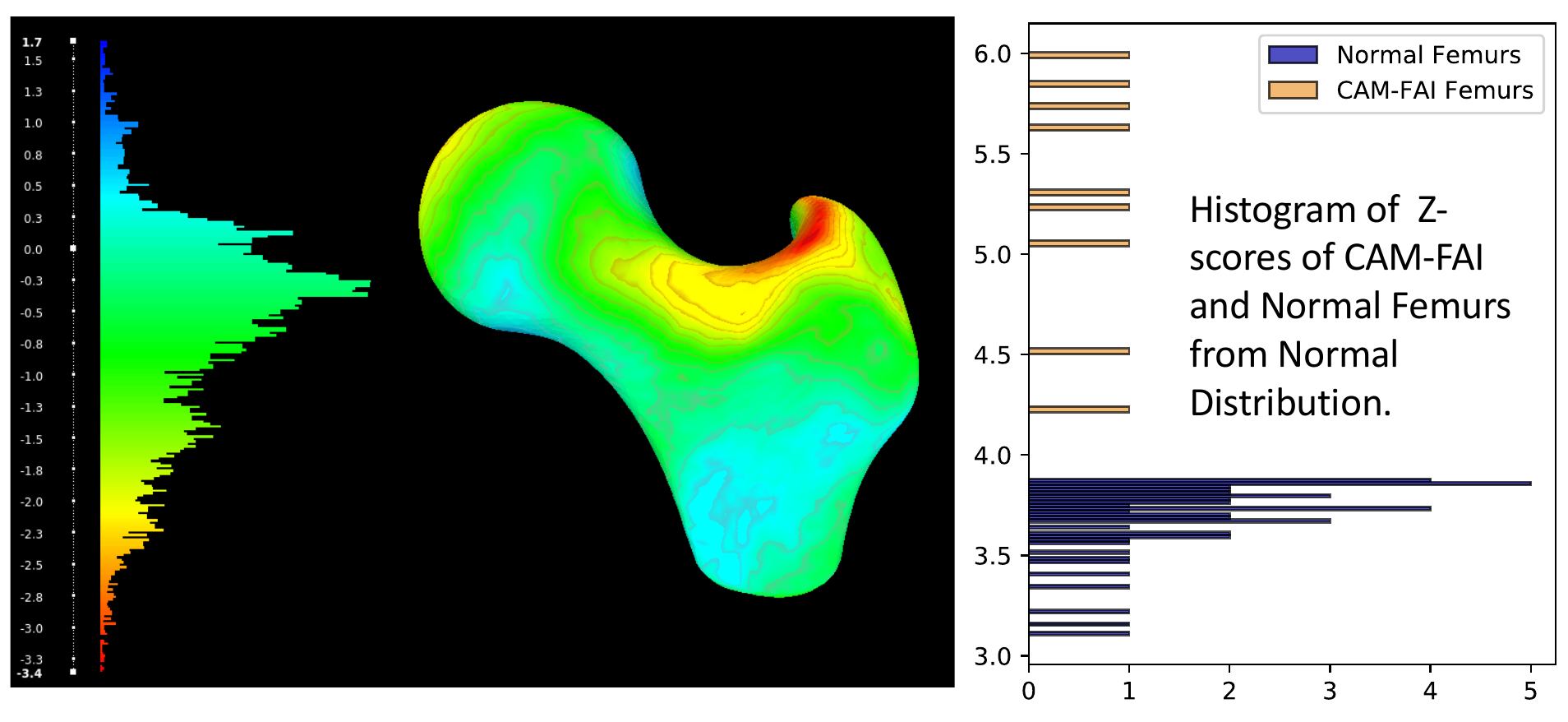}
    \caption{\textbf{CAM-FAI comparison with normal population: } (Left) shows the surface-to-surface difference between mean segmentation of normal mean shape to the mean segmentation of Cam-FAI mean shape, this is projected onto the mesh from normal mean shape. (Right) showcases the logarithm of Mahalanobis distance of each scan from the normal population using the landmarks as features.}
    \label{fig:compare}
\end{figure}

\textbf{Downstream Application: } We want to characterize how well is mean pathology showcased. In this experiment, we will use the segmentation masks of femur anatomy for each image available to us. We want to discover the mean anatomy of the femurs from the control set and mean of the femurs diagnosed with Cam-FAI, and we follow these steps for a given image set:
\begin{enumerate}
    \item We compute the mean landmarks (using the landmarks discovered by the proposed model) on a set of CT scans.
    \item We use these points to compute a mean image. We discover landmarks on each image, and then we register each segmentation mask to the mean landmarks, giving us a single approximation of the mean image.
    \item We perform such approximation for all the images in a population, and taking an average of all these approximations will give us the mean segmentation image.
\end{enumerate}
We follow the above steps for the control set, obtaining a representative shape for healthy femur anatomy (normal mean shape), and for the set of cam-FAI scans obtaining a representative femur shape with Cam-FAI (Cam-FAI mean shape). We compare the normal mean shape with Cam-FAI mean shape using the surface-to-surface distance from the Cam-FAI mean to normal; this distance is projected on the mesh of the normal mean shape in Figure \ref{fig:compare}. The negative values showcase the regions where average Cam-FAI pathology is outward, whereas positive values showcase it is inwards from the average normal. We see that visualized Cam-FAI pathology (outward regions) is similar to the clinically plausible location for the Cam-FAI-affected region on femur bone. We believe this behavior will be even more pronounced with more pathological scans in training. This experiment and the subsequent results are promising as the Cam-FAI deformity is very subtle and difficult to capture. It would require full segmentation and dense correspondences to capture such a subtle variation \cite{atkins2017quantitative}. Furthermore, we also compute the Mahalanobis distance using the landmarks as shape descriptors and normal CT forming the base distribution to verify whether CT scans with Cam-FAI diagnosis deviate from the population of controls. We clarify that the number of scans is less than the number of landmarks; we need to compute PCA using landmarks with dimension as the number of scans that captures 100\% of shape information. This is necessary; otherwise, the covariance is a singular matrix, and Mahalanobis distance will produce non-interpretable values. From the histogram shown in Figure \ref{fig:compare}, we see that the scans with Cam-FAI diagnosis significantly deviate from the control set, indicating that the shape descriptor is capturing the localized pathology well.

\subsection{Cardiac LGE Dataset}
\label{sec:cardiac-MRA}

This dataset consists of 3D late gadolinium enhancement (LGE) images of left-atrium (LA) for 207 patients. All these scans are from patients diagnosed with irregular heartbeats or atrial fibrillation (AF). These scans are acquired after the first ablation, a procedure used for the treatment of AF. Cardiac MR imaging was performed on AF patients presenting at the University of Utah Hospital’s Electrophysiology Clinic. Image sequences include a respiratory and ECG-gated MRA, acquired during continuous gadolinium contrast agent injection (0.1 mmol/kg, Multihance [Bracco Diagnostic Inc.]), followed by a 15-minute post-contrast LGE sequence.  Images were acquired on either a 1.5 T or 3 T clinical MR scanner (Siemens Medical Solutions) using phased-array receiver coils. LGE-MRI scans were acquired approximately 15 minutes after the contrast agent injection and were acquired at the end-diastole phase of the cardiac cycle. The scanning protocol utilized a 3D inversion recovery, respiration navigated, ECG-gated, gradient echo pulse sequence. Typical image acquisition parameters include the following: free-breathing using navigator gating, a transverse imaging volume with voxel size = 1.25 × 1.25 × 2.5 mm (reconstructed to 0.625 × 0.625 × 1.25 mm), and inversion time = 270–320 ms. Inversion times for the LGE-MRI scan were identified using a TI scout scan. Other parameters for the 1.5 T scanner included a repetition time of 5.4 ms, echo time of 2.3 ms, and a flip angle of 20°. Scans performed on the 3 T scanner were done using a repetition time of 3.1 ms, echo time of 1.4 ms, and a flip angle of 14°. ECG gating was used to acquire a small subset of phase encoding views during the diastolic phase of the LA cardiac cycle.

\begin{figure}
    \centering
    \includegraphics[width=\linewidth]{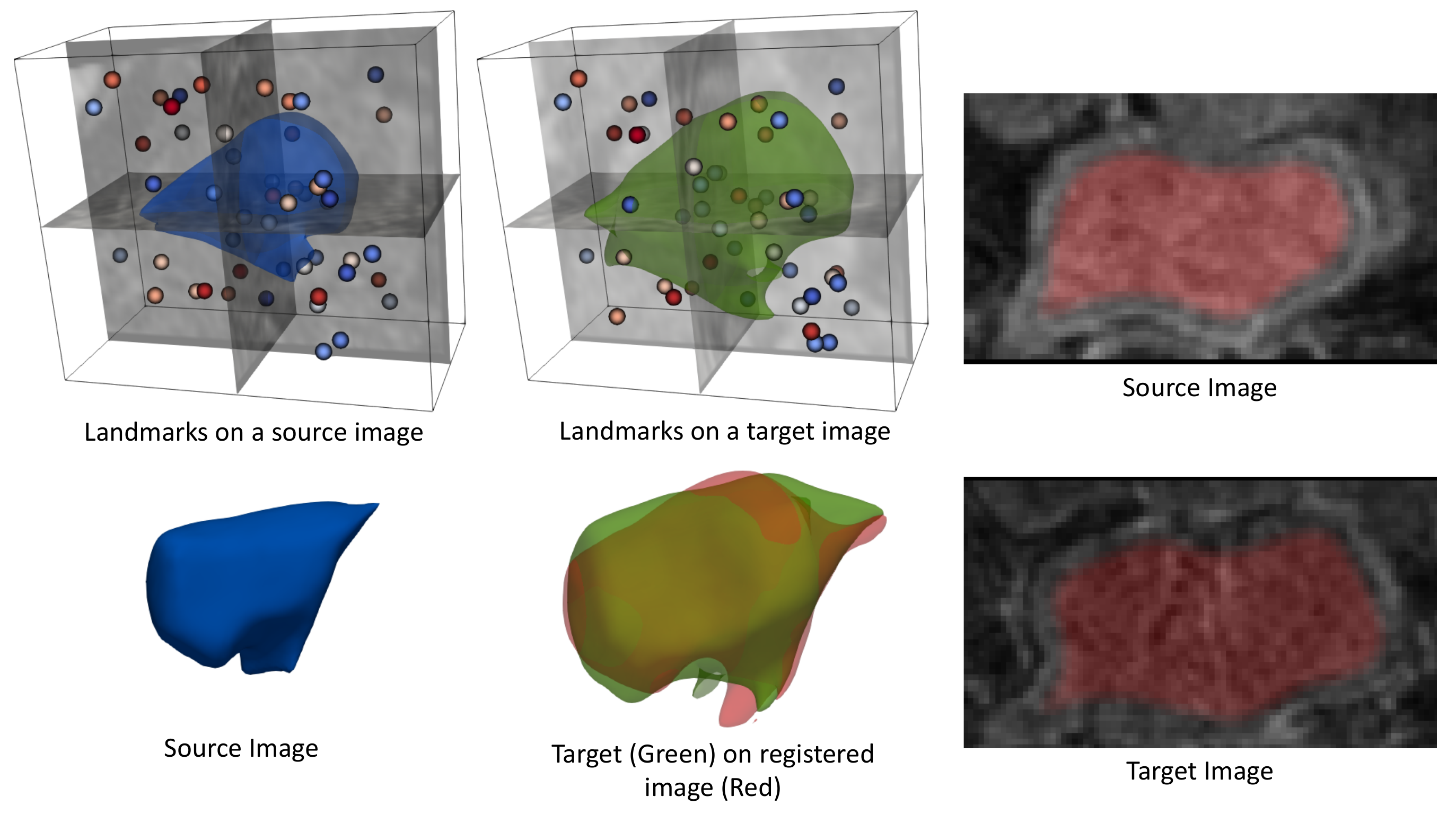}
    \caption{\textbf{Results on LGE Images} showcasing landmarks on a source image pair, and the registration of their segmentation. We also show mid axial slices of source and image to showcase the poor quality of the data.}
    \label{fig:la}
\end{figure}

It is a very challenging dataset due to two reasons, (i) these LGE images are low resolution and noisy, and very varied in terms of intensity profiles (see the image pair in Figure \ref{fig:la}), and (ii) the LA shape is not very distinguishable intensity-wise from the neighboring anatomies. We employ our model on this dataset to test its limits. The goal of this experiment is not to achieve accurate registration but rather to find usable shape descriptors. The downstream task for this application is the prediction of atrial fibrillation recurrence from LA shape, which is expressed via the proposed landmark shape descriptor. Due to the challenging nature of the data, we use the distance transforms via the weak supervision variant of the framework, and we use all the distance transforms (207 in number). We apply MIND loss for both the image matching terms on LGE and distance transforms, with a $\lambda=0.0000001$ and 100 landmarks. Additionally, we perform a center of mass alignment on the images (using the corresponding segmentations) followed by cropping. This step is necessary to highlight the LA shape and isolate it from the neighboring anatomies. We train the network for 15 epochs and perform redundancy removal by retaining 50 landmarks per image. A single source image pair with landmarks are shown in Figure \ref{fig:la}. The registration performance is not as good as other datasets described due to the low-quality images and high-variability nature of LGE intensities and LA shapes.

\textbf{Downstream Application: }   The LA scans are acquired post-ablation, however, even after ablation, AF can occur again, this is known as AF recurrence. Therefore, we also monitor patients post-ablation for a recurrence of AF. The shape of LA and the left-atrium appendage is shown to be successful in predicting AF recurrence \cite{bieging2018left} . For recurrence prediction, we again use a simple multi-layer perceptron with three hidden layers, which yields a test accuracy of $65.72\% \pm 1.32\%$. In comparison, we also use \emph{ShapeWorks} to place a dense set of 2048 particles on these left atrium shapes perform PCA on it to reduce to a 20-dimensional shape descriptor. Using this with the same MLP architecture for atrial fibrillation recurrence classification, we get a test accuracy of $68.34\% \pm 0.97\%$. This showcases that the discovered landmark shape descriptor captures almost the same amount of information captured by a dense correspondence model for this particular downstream task. A better initialization of landmarks that introduces a degree of supervision and domain expertise in the model will improve the shape descriptor. However, such experiments take away from the complete unsupervised and domain-independent framework of the model.



\section{Conclusions}
\label{sec:conclusions}

This paper proposes an end-to-end framework of obtaining usable \emph{shape descriptor} directly from a set of 2D/3D images. The proposed model is a self-supervised network that works under the assumption that anatomically consistent landmarks will register a pair of images well under a particular class of transformations. The model consists of a landmark encoder, an RBF solver, and a spatial transformer during training. For testing, we only use the landmark encoder to obtain a set of landmarks on a given image. The methodology has been initially proposed in our previous work \cite{bhalodia2020selfsupervised}. This paper provides detailed explanations and significantly extends it by introducing different image matching loss functions, two variants of loss functions that incorporate prior shape information, and extensive experimentation on several different 2D and 3D datasets. We find that the landmark shape descriptor obtained via the proposed model can be used directly for shape analysis and subsequent downstream tasks such as disease classification and severity quantification.
\section{Acknowledgements}

The National Institutes of Health supported this work under grant numbers NIBIB-U24EB029011, NIAMS-R01AR076120, NHLBI-R01HL135568, NIBIB-R01EB016701, NIBIB-R21EB026061, and NIGMS-P41GM103545. The content is solely the responsibility of the authors and does not necessarily represent the official views of the National Institutes of Health.  We would also like to thank Dr Jesse Goldstein, Dr Andrew Anderson, Dr Nassir Marrouche and Dr Penny Atkins for making their data available to be used in this manuscript.

\appendix
\section{RBF Solver for Image Registration}
\label{app:rbf}
If we consider the 2D image case now with $\mathbf{x} = (x,y)$ specifying a coordinate on the image grid. Let $T_x$ and $T_y$ be transformations acting on each coordinate, therefore the entire transformation is given as $T(\mathbf{x}) = (T_x(\mathbf{x}), T_y(\mathbf{y}))$. The individual transformations follow the RBF equation (Eq. \ref{eq:transform}), that results in:
\begin{align}
    T_x(\mathbf{x}) &= \sum\limits_{i=1}^M \phi(||\mathbf{x} - \mathbf{x}_i||)w_i^x + \alpha_2^x x + \alpha_1^x y +\alpha_0^x \\
    T_y(\mathbf{x}) &= \sum\limits_{i=1}^M \phi(||\mathbf{x} - \mathbf{x}_i||)w_i^y + \alpha_2^y x + \alpha_1^y y +\alpha_0^y
\end{align}

Here, $\mathbf{w}^x = (w_1^x, ..., w_M^x, \alpha_2^x, \alpha_1^x, \alpha_0^x)$ and $\mathbf{w}^y = (w_1^y, ..., w_M^y, \alpha_2^y, \alpha_1^y, \alpha_0^y)$ are the unknown parameters for x and y coordinates respectively. Now given a set of M control points on source and target image ($\mathbf{x}_i$ and $\bar{\mathbf{x}}_i$)we have $T(\mathbf{x}_i) = \bar{\mathbf{x}}_i = (\bar{x}_i, \bar{y}_i)$. These gives us 2M linear equations, we need six more equation to perfectly solve for $2(M+3)$ equations. We arrive at this via the constraint that the RBF part of the transformation should have no linear or constant term, i.e. $\sum\limits_{i=1}^M x_i w_i^x = 0$, $\sum\limits_{i=1}^M y_i w_i^x = 0$ and, $\sum\limits_{i=1}^M  w_i^x = 0$. We have three similar equation for y coordinate. There we have a system of equation given as:
\begin{align}
    \begin{bmatrix} B & 0 \\ 0 & B \end{bmatrix} \begin{bmatrix} \mathbf{w}^x  \\ \mathbf{w}^y \end{bmatrix} = \begin{bmatrix} \mathbf{k}_x \\ \mathbf{k}_y  \end{bmatrix} 
\end{align}

Where, $\mathbf{k}_x = [0, 0, 0, \bar{x}_1, ..., \bar{x}_M]$ and $\mathbf{k}_y = [0, 0, 0, \bar{y}_1, ..., \bar{y}_M]$, and 
\begin{align}
    B = \begin{bmatrix}
x_{1}       & x_{2}  & \dots & x_{M} & 0 & 0 & 0 \\
y_{1}       & y_{2} & \dots & y_{M} & 0 & 0 & 0 \\
1       & 1 & \dots & 1 & 0 & 0 & 0 \\
\phi_{11} & \phi_{12} & \dots & \phi_{1M} & x_1 & y_1 & 1\\
\phi_{21} & \phi_{22} & \dots & \phi_{2M} & x_2 & y_2 & 1\\
\hdotsfor{7} \\
\phi_{M1} & \phi_{M2} & \dots & \phi_{MM} & x_M & y_M & 1\\
\end{bmatrix}
\end{align}

Now for solving the transformation parameters we just solve the following linear equation with $A\mathbf{w} = b$, where $A = \begin{bmatrix} B & 0 \\ 0 & B \end{bmatrix}$ and $b = \begin{bmatrix} \mathbf{k}_x \\ \mathbf{k}_y  \end{bmatrix}$.

\section{Architectural Details}
\label{app:arch}
\begin{figure}
    \centering
    \includegraphics[width=\linewidth]{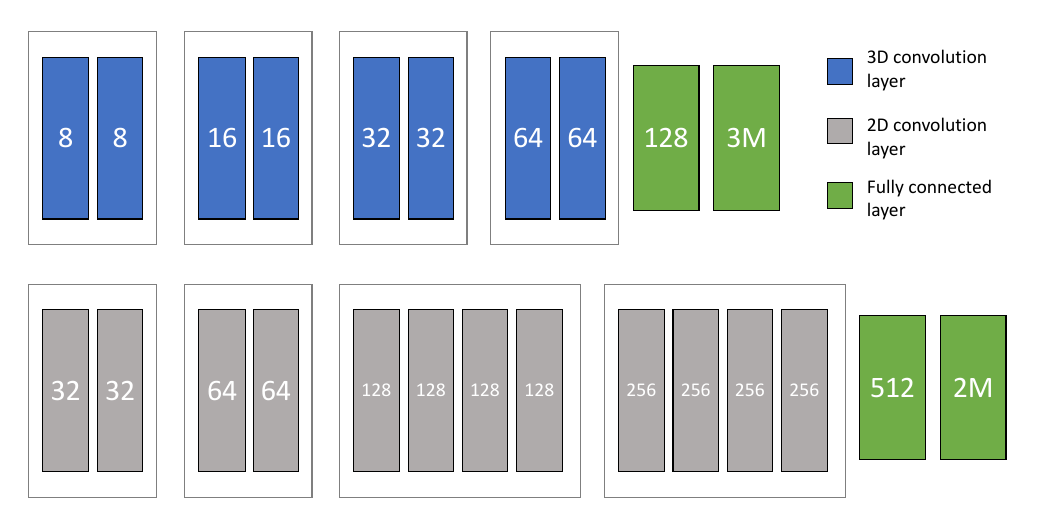}
    \caption{\textbf{Detailed Network Architectures} (Top) For 3D data, (Bottom) for 2D data.}
    \label{fig:suparch}
\end{figure}

The architecture of landmark encoder used for all 2D experiments is given by bottom Figure \ref{fig:suparch}, it is comprised of four blocks, first two blocks having two convolution layers and the last two having four. Each block is followed by a max pooling with factor 2. Each convolutional layer has a kernel size of $3 \times 3$. The activation function used throughout is ReLU, except at the output layer we use hyperbolic tangent. The output points are in the range -1 to 1 which are then scaled to the original coordinates by using the image dimensions.

For 3D we have a smaller architecture due to memory constraints, it consists of four blocks with two convolutional layers each. Each block is followed by a max pooling with factor 2. Each convolutional layer has a kernel size of $3 \times 3 \times 3$. The activation function used here is a leaky-ReLU except at the output layer where we use hyperbolic tangent.

\bibliographystyle{splncs04}
\bibliography{references}

\end{document}